# A Landscape-Aware Differential Evolution for Multimodal Optimization Problems

Guo-Yun Lin, *Student Member*, *IEEE*, Zong-Gan Chen, *Member*, *IEEE*, Chuanbin Liu, Yuncheng Jiang, *Member*, *IEEE*, Sam Kwong, *Fellow*, *IEEE*, Jun Zhang, *Fellow*, *IEEE*, and Zhi-Hui Zhan, *Fellow*, *IEEE*

*Abstract*—How to simultaneously locate multiple global peaks and achieve certain accuracy on the found peaks are two key challenges in solving multimodal optimization problems (MMOPs). In this paper, a landscape-aware differential evolution (LADE) algorithm is proposed for MMOPs, which utilizes landscape knowledge to maintain sufficient diversity and provide efficient search guidance. In detail, the landscape knowledge is efficiently utilized in the following three aspects. First, a landscape-aware peak exploration helps each individual evolve adaptively to locate a peak and simulates the regions of the found peaks according to search history to avoid an individual locating a found peak. Second, a landscape-aware peak distinction distinguishes whether an individual locates a new global peak, a new local peak, or a found peak. Accuracy refinement can thus only be conducted on the global peaks to enhance the search efficiency. Third, a landscape-aware reinitialization specifies the initial position of an individual adaptively according to the distribution of the found peaks, which helps explore more peaks. The experiments are conducted on 20 widely-used benchmark MMOPs. Experimental results show that LADE obtains generally better or competitive performance compared with seven well-performed algorithms proposed recently and four winner algorithms in the IEEE CEC competitions for multimodal optimization.

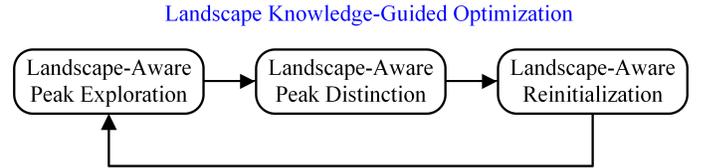

**Fig. 1.** Basic framework of LADE.

## I. INTRODUCTION

OPTIMIZATION problems with multiple global optima are known as multimodal optimization problems (MMOPs). Considering the shape of the fitness landscape in MMOPs, an optimum is also termed a peak. MMOPs arise in various real-world fields, and have been a popular research topic in recent years [1]–[4].

Evolutionary computation (EC) algorithms have shown promising performance in solving various optimization problems [5]–[8]. Since conventional EC algorithms are designed to locate a single global optimum, how to maintain sufficient search diversity to simultaneously locate multiple peaks in MMOPs is a key challenge in designing EC-based multimodal optimization algorithms. Niching is the most popular technique, whose principle is to divide the entire population into multiple subpopulations and those individuals of the same subpopulation are supposed to be in the region of the same peak [9]–[12]. In this way, each subpopulation evolves independently and acts as a unit to locate a peak. Thus, multiple peaks can be located by multiple subpopulations. To further enhance the search efficiency, a distributed individuals for multiple peaks (DIMP) framework is proposed in [13], in which each individual acts as a unit to locate a peak with the help of a virtual population.

However, existing approaches do not sufficiently utilize landscape knowledge in solving MMOPs. During the search process, algorithms will gradually capture the landscape of MMOPs, which contains useful landscape knowledge to help maintain sufficient diversity and provide efficient search guidance. First, the knowledge of found peaks is not utilized to avoid duplicate search efforts on the same peak. Thus, multiple individuals may locate the same peak, particularly in the case that a peak has a very large region. Second, there is no distinction of the found peaks, i.e., distinguish whether a found peak is a global or local peak. As a result, search effort may be persistently wasted on refining the solution accuracy on a local peak. Third, the individuals are usually initialized randomly in the search space without considering landscape knowledge. However, since there are multiple peaks in the search space and an individual is more likely to locate a peak in its vicinity, landscape knowledge can help achieve an efficient initialization to help locate more peaks.

To efficiently utilize the landscape knowledge to guide the optimization, a landscape-aware differential evolution (LADE)

This work was supported in part by the National Natural Science Foundation of China under Grant 62206100, Grant 62176094, and Grant U23B2039, in part by the Guangdong Basic and Applied Basic Research Foundation under Grant 2024A1515011708, in part by the Guangzhou Basic and Applied Basic Research Foundation under Grant 2023A04J0319, in part by the Tianjin Top Scientist Studio Project under Grant 24JRRCRC00030, in part by the Tianjin Belt and Road Joint Laboratory under Grant 24PTLYHZ00250, in part by the Fundamental Research Funds for the Central Universities, Nankai University (078-63243159, 078-63241453, and 078-63243198), and in part by the research fund of Hanyang University (HY-202300000003465 and HY-202400000001955). (*Corresponding authors: Zong-Gan Chen; Chuanbin Liu.*)

Guo-Yun Lin, Zong-Gan Chen, and Yuncheng Jiang are with the School of Computer Science, South China Normal University, Guangzhou 510631, China. (e-mail: charleszg@qq.com)

Chuanbin Liu is with the Center for Scientific Research and Development in Higher Education Institutes, Ministry of Education, P.R.China.

Sam Kwong is with the Department of Computing and Decision Science, Lingnan University, Hong Kong.

Jun Zhang is with the College of Artificial Intelligence, Nankai University, Tianjin 300350, China, and also with the Hanyang University, ERICA, 15588, South Korea.

Zhan-Hui Zhan is with the College of Artificial Intelligence, Nankai University, Tianjin 300350, China.

algorithm for MMOPs is proposed in this paper. The basic framework of LADE is shown in Fig. 1, in which the landscape knowledge is efficiently utilized in three aspects, i.e., landscape-aware peak exploration, landscape-aware peak distinction, and landscape-aware reinitialization.

Landscape-aware peak exploration aims to explore multiple peaks efficiently in the search space. To achieve this goal, individuals evolve adaptively according to their corresponding virtual population and lifetime. The previous work [13] has validated that such an evolutionary scheme can help each individual locate a peak. To further enhance the search efficiency, landscape knowledge is utilized to prevent an individual from locating a found peak. Specifically, landscape-aware peak exploration analyzes the search history to capture the shape of the found peaks and simulates the regions of them. Then, the simulated peak regions are treated as taboo regions in the following search of individuals to encourage individuals to explore new peaks.

Since individuals have located multiple peaks according to the landscape-aware peak exploration, landscape-aware peak distinction aims to distinguish whether a found peak is a global or local peak, helping obtain more landscape knowledge. Particularly, obtained solutions on the found peaks may still not meet the accuracy requirement. With the help of landscape-aware peak distinction, LADE can only refine the solution accuracy on global peaks to save function evaluations (FEs). The fitness improvement rate during the evolutionary process and the fitness gap of the individual toward the historical best fitness are utilized to distinguish whether the individual locates a new global peak. Then, a hill-valley test is conducted to further distinguish whether the individual locates a new local peak. In addition, a local search strategy is proposed to refine the solution accuracy on the global peaks according to an adaptive scheme.

After the distinction of the peak found by the individual, the individual needs to be reinitialized and then starts a new peak exploration process. To reinitialize the individual in a region with unexplored peaks and help locate more peaks, landscape-aware reinitialization specifies the initial position and search range of the individual based on the distribution and distinction of found peaks. On the one hand, landscape-aware reinitialization clusters the found peaks and analyzes the fitness values of peaks in each cluster to detect whether there may exist a potential global peak surrounded by the found peaks in a cluster. If so, an individual will be reinitialized in a region constructed based on the distribution of found peaks in the corresponding cluster to help locate a potential global peak. On the other hand, allocating more search effort to the region with fewer found peaks is beneficial to maintain sufficient diversity and help explore new peaks. Thus, based on the distribution of the found global peaks, landscape-aware reinitialization recursively divides the entire search space into multiple subspaces. An individual will be reinitialized in one of the subspaces according to a probability distribution and the search range of the individual is also restricted in the selected subspace. The fewer global peaks in a subspace, the higher the probability that the subspace is selected.

By utilizing the landscape knowledge to achieve efficient peak exploration, peak distinction, and reinitialization, the proposed LADE can maintain sufficient search diversity and provide efficient search guidance in solving MMOPs. To validate the effectiveness of LADE, the benchmark set used in the latest IEEE CEC competition for multimodal optimization [14] is adopted, which is also widely used to evaluate the algorithms for MMOPs recently [15]–[19]. 20 MMOPs with various features are included in the benchmark set. Compared with 11 state-of-the-art and well-performed multimodal optimization algorithms, including four winner algorithms in the IEEE CEC competitions for multimodal optimization, LADE obtains generally better or competitive performance.

The rest of the paper is organized as follows: Section II introduces the differential evolution (DE) and multimodal optimization techniques in the literature; Section III introduces the LADE algorithm and its novelty in detail; Section IV presents the experimental studies; Finally, Section V draws a conclusion.

## II. RELATED WORK

### A. Differential Evolution

DE is an EC algorithm based on difference vectors among individuals [20]–[23]. In a conventional DE [24], a population is randomly initialized at first. Then, the population is updated according to the mutation, crossover, and selection operations iteratively until the termination criterion is met. Details of the three main operations are presented as follows.

*Mutation*: Each individual $X_i$ in the population generates a mutant $V_i$ according to (1), where $X_{r1}$, $X_{r2}$, and $X_{r3}$ are the other three individuals randomly selected from the population. $F$ is a parameter named scaling factor.

$$V_i = X_{r1} + F \times (X_{r2} - X_{r3}) \qquad (1)$$

*Crossover*: The crossover operation is conducted based on $X_i$ and $V_i$, as shown in (2). An offspring of $X_i$ (i.e., $U_i$) is generated by selecting the value of $X_i$ or $V_i$ in each dimension according to a certain crossover rate $CR$. To ensure that $U_i$ is different from $X_i$ in at least one dimension, $d_{rand}$ is an integer randomly chosen from the range $[1, D]$, where $D$ is the dimension size of the optimization problem.

$$U_i^d = \begin{cases} V_i^d, & \text{if } \text{rand}(0,1) \le CR \text{ or } d = d_{rand} \\ X_i^d, & \text{otherwise} \end{cases} \qquad (2)$$

*Selection*: As shown in (3), if the offspring $U_i$ has a better fitness value than $X_i$, DE will replace $X_i$ with $U_i$ for the next generation. Otherwise, $X_i$ remains unchanged.

$$X_i = \begin{cases} U_i, & \text{if } f(U_i) \ge f(X_i) \text{/*for maximization problems*/} \\ X_i, & \text{otherwise} \end{cases} \qquad (3)$$

### B. Multimodal Optimization Techniques

To enhance the capability of EC algorithms in solving MMOPs, various techniques have been proposed in recent years, which can mainly be classified into the following three categories.

*1) Niching*

In niching methods, the population is divided into multiple subpopulations to maintain diversity. Crowding [25], [26] and speciation [27] are two classical niching methods. The crowding method lets an offspring compete with the most similar individual in the subpopulation. The speciation method conducts population division based on the species radius, allowing independent evolution among different subpopulations. However, additional parameters required in these two methods, such as the subpopulation size in the crowding method and the species radius in the speciation method, are sensitive and difficult to tune.

To achieve a better population division, clustering methods are incorporated into the algorithms [28], [29]. Luo *et al.* [10] used the nearest-better clustering to divide the population roughly, and then adaptively merge the neighboring subpopulations according to the fitness difference between subpopulations. Fieldsend [30] proposed a niching migratory multiswarm optimizer, which merged two subpopulations if they were found to be exploring the same peak. Li *et al.* [31] detected whether two individuals were in the region of the same peak according to a hill-valley test based on historical individuals. To assist the population division, the network community-based differential evolution for MMOPs [17] constructs the population as an attribute network based on the search history of individuals and the differences among individuals.

*2) New Evolutionary Operator*

Many researchers have designed new evolutionary operators to enhance the performance of EC algorithms in solving MMOPs. Lin *et al.* [32] proposed two keypoint-based mutation operators to deal with the situation where a subpopulation contained individuals locating different peaks. The two new mutation operators cooperate with two classical mutation operators to provide efficient guidance. Sun *et al.* [33] proposed two new mutation operators that incorporated the best individual in each subpopulation to accelerate the convergence on peaks. Liao *et al.* [34] designed a new mutation operator by combining historical offspring that did not obtain improvement. The adaptive differential evolution with archive [35] samples individuals between the current individual and its nearest better individual. The sampled individuals are used to generate mutant individuals.

*3) Multiobjective Transformation*

Some researchers have transformed MMOPs into multiobjective optimization problems. A popular approach is to set the fitness value as the first optimization objective and the second objective is a diversity metric to help maintain population diversity [36]. Cheng [37] mapped the individuals into a grid coordinate system to evaluate the population diversity. Wang *et al.* [38] formulated a bi-objective optimization problem in each dimension, which could transform peaks in MMOPs into the Pareto optimal solutions in the formulated multiobjective optimization problems.

Although various techniques have been proposed to enhance the performance of EC algorithms on MMOPs, only a few researchers have made attempts to utilize the landscape knowledge that has great potential. Wei *et al.* [15] designed a dynamic penalty strategy, which penalized the fitness of individuals near the found peaks. Ahrari *et al.* [15], [39] adaptively estimated the regions of peaks. However, landscape knowledge is still not utilized efficiently. To enhance search diversity and provide efficient search guidance, LADE is proposed in this paper, which utilizes landscape knowledge to guide the peak exploration, peak distinction, and reinitialization processes.

III. LANDSCAPE-AWARE DIFFERENTIAL EVOLUTION

LADE utilizes landscape knowledge included in search trajectories to maintain sufficient diversity and enhance search efficiency. First, in the landscape-aware peak exploration, peak exploration procedure (PEP) lets each individual sufficiently explore the search space to locate a peak. Peak region simulation strategy (PRSS) simulates the regions of the found peaks, which can help individuals avoid locating the found peaks during PEP. Second, in the landscape-aware peak distinction, peak distinction mechanism (PDM) is proposed to distinguish whether an individual locates a new global peak, a new local peak, or a found peak. Note that a peak list ***P*** stores solutions found on the global and local peaks, and a global peak list ***GP*** includes solutions found on global peaks. In addition, the local search strategy (LSS) is incorporated to refine the solution accuracy on the found global peaks. Third, the landscape-aware reinitialization adaptively reinitializes an individual to help locate more peaks. Specifically, potential optimal region detection strategy (PORDS) reinitializes an individual in a region where a potential global peak is surrounded by multiple found peaks, while subspace division strategy (SDS) reinitializes an individual in a region where fewer global peaks are found.

Note that individuals in LADE are mapped into a normalized search space in each dimension. Since different MMOPs have different search ranges, normalization can reduce the difficulty of parameter configurations and enhance the robustness of LADE.

*A. Landscape-Aware Peak Exploration*

During the landscape-aware peak exploration, PEP lets an individual act as a distributed unit to locate a peak in the search space. When an individual locates a peak, the search trajectories of individuals that contain landscape knowledge are utilized by PRSS to simulate the peak region. In the subsequent search, an individual will keep away from the simulated peak regions to help locate a new peak. Details of PEP and PRSS are presented as follows.

*1) Peak Exploration Procedure*

Based on the DIMP framework [13], PEP lets an individual serve as a distributed unit to locate a peak. Specifically, an individual ***X*** conducts mutation operation based on its virtual population according to (4). ***VX***$_1$ and ***VX***$_2$ are two virtual individuals corresponding to ***X*** and are generated by (5), where $d \in \{1, 2, \ldots, D\}$ and $R$ controls the range of virtual individuals. Note that $LB^d$ and $UB^d$ are the lower bound and the upper bound of the normalized search space in the $d$-th dimension. After that, the conventional crossover operation shown in (2) is conducted to obtain an offspring $U$. Note that to avoid an individual locating a found peak, mutation and crossover operations are conducted iteratively until the generated offspring is not located in the regions of found peaks simulated by PRSS.

$$V = X + F \times (VX_1 - VX_2) \quad (4)$$

$$VX^d = \text{rand}(\max(X^d - R/2, LB^d), \min(X^d + R/2, UB^d)) \quad (5)$$

In order to help an individual sufficiently explore the search space and gradually converge to a peak, $R$ is initialized to a larger value (i.e., 1) in the normalized search space and adaptively decreases according to the evolutionary status of the individual. In detail, if the consecutive generations that the individual fails to improve its fitness in the selection operation reach the maximal consecutive generations $mcg$, the value of $R$ will be halved. Moreover, a lifetime threshold $lt$ is incorporated. If the $R$ of an individual has decreased $lt$ times, the lifetime of the individual is exhausted and its corresponding solution is then sent to PDM to distinguish whether it locates a global or local peak. After that, the individual is reinitialized and starts a new lifetime. Due to the page limit, the complete procedure of PEP is shown in Algorithm S.1 in the supplementary material. Note that compared with the DIMP framework in [13], PEP further considers the simulated peak regions during the search of an individual to prevent the individual from locating a found peak, helping enhance the search efficiency.

*2) Peak Region Simulation Strategy*

Individuals can locate peaks according to PEP, but they may find the same peaks without a constraint on their search range. To avoid wasting FEs caused by locating the same peak, LADE incorporates PRSS to simulate the regions of found peaks and lets the simulated peak regions act as taboo regions in PEP. Generally, a peak region can be formulated as a search space centered around the peak and with a certain boundary. Fig. 2 shows examples of a peak region in 1-D and 2-D MMOPs. Each triangle in Fig. 2 represents a peak. For a peak $P_k$, a vector termed $PD_k$ is used to specify the range of the peak region, which represents the distance between $P_k$ and the boundary of the peak region in each dimension.

However, there are various peaks with different features, how to determine the boundary of a peak region is thus a challenging issue. Motivated by the hill-valley landscape of peaks, PRSS searches for historical individuals who are close to the peak and have worse fitness. Then, a region centered around the peak located by an individual is determined according to the distribution of the historical individuals. The detailed procedure of PRSS is shown in Algorithm 1. Note that the position of the individual who first locates $P_k$ is used to represent $P_k$ and the set $S$ is used to store the individuals in the region of $P_k$. First, $P_k$ is added to $S$ (Step 1). Then, Steps 3–6 recursively search from all historical individuals to find those in the region of $P_k$. In detail, for each individual $S_{cur}$ in $S$, all subordinate neighbors of $S_{cur}$ are added to $S$. Such a recursive process finishes until no new subordinate neighbor is

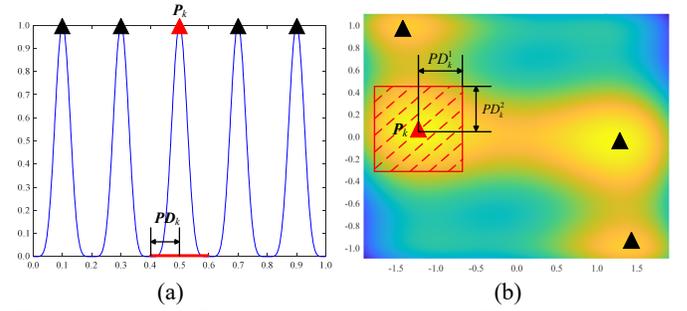

**Fig. 2.** Examples of simulated peak regions. (a) The red line segment represents the peak region in a 1-D MMOP. (b) The red rectangle represents the peak region in a 2-D MMOP.

---

**Algorithm 1** PRSS
1. Add $P_k$ to an empty set $S$.
2. $cur = 1$;
3. **While** $cur \leq |S|$
4.     Add all subordinate neighbors of $S_{cur}$ to $S$;
5.     $cur = cur + 1$;
6. **End While**
7. $MD^d = \max\limits_{m \in \{1,2,...,|S|\}} \left( | P_k^d - S_m^d | \right), d \in \{1, 2, ..., D\}$;
8. **If** $\prod_{d=1}^{D} MD^d > \prod_{d=1}^{D} (\mu \times PD_k^d)$
9.     $PD_k = MD$;
10. **Else**
11.     $PD_k = \mu \times \sqrt[D]{\prod_{d=1}^{D} \frac{PD_k^d}{MD^d}} \times MD$ ;
12. **End If**

---

found. Specifically, a historical individual is the subordinate neighbor of $S_{cur}$ if $S_{cur}$ has a better fitness value and the Euclidean distance between them is no more than the simulation distance $sd$. Considering that the larger the search space, the sparser the search trajectories, $sd$ is set adaptively according to (6) based on the dimension size of the MMOP. On low-dimensional MMOPs, $sd$ is set to a small value to help rigorously determine the peak boundary. On high-dimensional MMOPs, the distribution of historical individuals is usually sparse, and a large $sd$ is thus set to avoid missing the historical individuals belonging to the same peaks.

$$sd = 0.005 \times \left( \lfloor D/5 \rfloor + 1 \right) \quad (6)$$

After constructing $S$, PRSS calculates the maximal distance $MD$ between $P_k$ and individuals in $S$ in each dimension (Step 7). Lastly, $PD_k$ is set based on $MD$ according to Steps 8–12. Note that with limited search trajectories, it is very difficult to capture the complete peak region the first time the peak is located. Thus, a peak will still be located multiple times by individuals. To gradually obtain a complete peak region, PRSS is conducted every time the peak is located. Considering the individual who is not the first one locating the peak usually lies right next to the boundary of the peak region, PRSS always starts the simulation from the position of the individual who first locates the peak. Moreover, to ensure the peak region is enlarged sufficiently, a parameter $\mu$ is incorporated, which guarantees that the peak region is enlarged a least $\mu$ times than the previous one. To adapt to MMOPs of different scales, $\mu$ is set to $(1.15 + 0.1 \times \lfloor D/5 \rfloor)$.

*B. Landscape-Aware Peak Distinction*

The search trajectories of individuals contain a lot of knowledge about the landscape of the search space. PDM analyzes the search trajectories to distinguish the detailed situation of a lifetime-exhausted individual. Locating a new global peak and locating a new local peak are the two most important situations that LADE focuses on. In addition, LSS is conducted on the found global peaks to refine the solution accuracy. Meanwhile, LSS also helps correct a few misclassifications of PDM in solving very complex MMOPs.

*1) Peak Distinction Mechanism*

PDM is proposed to distinguish the feature of the peak located by a lifetime-exhausted individual. Two indicators termed scaling fitness distance *SFD* and fitness improvement rate *FIR* are adopted. *SFD* of a lifetime-exhausted individual $X$ is calculated by (7), where $\lambda$ is a parameter ranging in [0,1] and $X_{best}$ represents the historical individual with the best fitness. *FIR* calculates the average fitness improvement of $X$ during the *tg* generations before the *lg*-th generation, as shown in (8). Note that according to PEP, a lifetime-exhausted individual has no fitness improvement during the last *mcg* generation. To obtain a useful indicator, *FIR* starts from the *mcg*-th generation before the last generation to calculate the average fitness improvement. In detail, suppose that $G$ represents the number of the last generation of $X$, *lg* equals $(G-mcg)$. $X_{lg}$ thus represents the historical individual of $X$ obtained in the *mcg*-th generation before the last generation.

$$SFD(X) = \lambda \times |f(X_{best}) - f(X)| \tag{7}$$

$$FIR(X) = \frac{|f(X_{lg}) - f(X_{lg-tg})|}{tg} \tag{8}$$

Different global peaks have different fitness landscapes. Large-region, gently sloping peaks and small-region, steeply sloping peaks are two typical types. Large-region, gently sloping peaks are easier for individuals to locate. Thus, a lifetime-exhausted individual who locates such a peak can achieve sufficient convergence on the peak quickly and obtain a good fitness. In this case, the values of *SFD* and *FIR* are both small. Oppositely, a lifetime-exhausted individual who locates a global peak with a small region and steep slope may still struggle to improve its fitness to reach the peak. Therefore, the values of *SFD* and *FIR* are both large. Based on the above observation, the relationship between *SFD* and *FIR* can help distinguish whether a lifetime-exhausted individual locates a new global peak. In addition, the hill-valley test [40] is adopted to help further distinguish whether a new local peak is located.

The detailed procedure of PDM is shown in Algorithm 2. First, for a lifetime-exhausted individual $X$, if its *SFD* is smaller than its *FIR* (Step 1), the individual is estimated to locate a new global peak and will be added to $P$ and $GP$ (Step 2). $\lambda$ in (7) acts as a scaling coefficient to help achieve an accurate distinction, and a comprehensive parameter analysis of $\lambda$ is presented in Section IV-D. In (8), *tg* is set adaptively based on the dimension size of the MMOP, as shown in (9). When solving a MMOP with a higher dimension size, an individual usually needs more generations to explore a peak, a larger *tg* is thus more appropriate to measure the fitness improvement rate.

**Algorithm 2** PDM
1. **If** $SFD(X) < FIR(X)$
2.    Add $X$ to $P$ and $GP$;
3. **Else**
4.    $P_k = \arg\min\limits_{P_j \in P}(PRRD_j)$;
5.    **If** hill-valley($X$, $P_k$, *hvnum*) = FALSE
6.      Add $X$ to $P$;
7.    **End If**
8. **End If**

$$tg = 80 \times 2^{\lfloor D/10 \rfloor + 1} \tag{9}$$

Then, if $X$ is not estimated to locate a new global peak, PDM further distinguishes whether $X$ locates a new local peak (Steps 4–7). PDM finds one of the found peaks (termed $P_k$) that is closest to $X$ (Step 4). Specifically, a peak region relative distance (termed *PRRD*) is defined in (10) to measure the distance between $X$ and a found peak $P_j$. The farther between $X$ and $P_j$, the larger $PRRD_j$ is. Note that $X$ is not allowed to enter the peak regions, thus *PD* of a peak region in each dimension is used as the denominator in (10) to achieve a fair distance measure among multiple found peaks. After that, the hill-valley test is performed between $X$ and $P_k$ with the sampling times *hvnum* to check if $X$ and $P_k$ locate the same peak (Step 5). The larger the search space, the more complicated the fitness landscape is. More sampling times are needed to detect the situation between $X$ and $P_k$. Thus, *hvnum* is set adaptively according to (11). If the hill-valley test obtains the result "FALSE", which represents $X$ and $P_k$ do not locate the same peak, $X$ is added to $P$ as a new local peak (Step 6).

$$PRRD_j = \sqrt{\sum_{d=1}^{D}(\frac{X^d - P_j^d}{PD_j^d})^2} \tag{10}$$

$$hvnum = 10 + 2 \times D \tag{11}$$

*2) Local Search Strategy*

LSS aims to refine the solution accuracy of global peaks in *GP*. Meanwhile, LSS also helps correct a few misclassifications of PDM in solving very complex MMOPs. The procedure of LSS is shown in Algorithm 3. Sampling new solutions around the found peaks is promising to achieve better solution accuracy. With limited FEs and multiple peaks, conducting the local search on each peak equally may not yield good overall performance. To determine which peaks to conduct the local search and how many new solutions are sampled, the local search probability (termed *LSP*) of each peak and the number of sampled solutions (termed *snum*) should be calculated at first (Step 1). $LSP_i$ of $GP_i$ is calculated by (12). If $LSP_i$ equals 1, LSS conducts the local search around $GP_i$ (Step 2). The setting of *LSP* assigns a higher probability to a peak whose fitness is worse, which efficiently allocates FEs on different peaks and helps achieve satisfying accuracy on each peak. *snum* is adaptively calculated in (13). The higher the dimension size, the more sampled solutions are needed. In addition, *opnum* represents the number of peaks whose *LSP* is 1. A larger value of *opnum* indicates more peaks in *GP* that may still not meet the accuracy requirement, and thus a smaller *snum* is set to avoid allocating too many FEs on local search.

$$LSP_i = \begin{cases} 1, & \text{if rand}(0,1) \leq \dfrac{1}{1+e^{20-2\times 10^7 \times |f(X_{best})-f(GP_i)|}} \\ 0, & \text{otherwise} \end{cases} \quad (12)$$

$$snum = \left\lceil 3 \times D \times \min\left(\dfrac{|GP|}{opnum}, 10\right) \right\rceil \quad (13)$$

In LSS, a sampled solution $SS_i$ around a peak $GP_i$ is generated in Step 4 based on the Gaussian distribution, in which $GP_i$ is the mean and $\sigma_i$ is the standard deviation. The setting of $\sigma_i$ is essential to help achieve efficient accuracy refinement. A fixed value cannot obtain satisfying performance on various MMOPs. Thus, $\sigma_i$ is set adaptively, and each peak $GP_i$ has its independent $\sigma_i$. First, $\sigma_i$ is initialized to $\sigma_{ini}$ when $GP_i$ is added to *GP*. A stagnation counter $sc_i$ is used to record the consecutive times that $SS_i$ fails to achieve a better fitness to replace $GP_i$ (Step 5). If $sc_i$ is larger than a descent threshold *dt*, reset $sc_i$ to 0 and divide $\sigma_i$ by 5 (Steps 6–8). When $\sigma_i$ is not larger than $\sigma_{ter}$, a complete local search process has been performed on $GP_i$. Then, $\sigma_i$ is reset to $\sigma_{ini}$ and $sc_i$ is initialized to 0 (Step 10), which lets $GP_i$ start a new local search process. Note that $lsnum_i$ records the number of local search processes conducted on $GP_i$.

However, a few local peaks may be misclassified as global peaks. Such a peak cannot achieve a fitness value the same as a real global peak no matter how many local search processes are conducted. To avoid useless local search processes and save FEs, a further distinction is conducted to determine whether to keep $GP_i$ in *GP* (Step 11). One type of peaks that may be misclassified is local peaks with extremely steep slopes, since *FIR* may be very large in this case. To deal with this problem, the fitness gap rate $FGR_i$ is calculated in (14), where $X_{worst}$ represents the historical worst individual. If $FGR_i \times \sqrt{lsnum_i} > 0.04$, $GP_i$ is considered as a local peak and will be removed from *GP*. A worse fitness value of $GP_i$ leads to a higher $FGR_i$. In addition, if $GP_i$ cannot achieve high accuracy during multiple local search processes, $GP_i$ is more likely to be a local peak. Thus, $lsnum_i$ is also considered in the distinction.

$$FGR_i = \dfrac{f(X_{best}) - f(GP_i)}{f(X_{best}) - f(X_{worst})} \quad (14)$$

**Algorithm 3** LSS
1. Calculate $LSP_i$ ($i \in \{1, 2, ..., |GP|\}$) and *snum*;
2. **For** each global peak $GP_i$ satisfying $LSP_i = 1$
3.    **For** $s = 1$: *snum*
4.       $SS_i$ = Gaussian($GP_i$, $\sigma_i$);
5.       Replace $GP_i$ with $SS_i$ and reset $sc_i$ to 0 if $SS_i$ is better than $GP_i$, otherwise increase $sc_i$ by 1;
6.       **If** $sc_i > dt$
7.          **If** $\sigma_i > \sigma_{ter}$
8.             $sc_i = 0$, $\sigma_i = \sigma_i / 5$;
9.          **Else**
10.            $sc_i = 0$, $\sigma_i = \sigma_{ini}$, $lsnum_i = lsnum_i + 1$;
11.            Determine whether to keep $GP_i$ in *GP*;
12.          **End If**
13.       **End If**
14.    **End For**
15. **End For**
16. Cluster global peaks in *GP* by mean-shift clustering;
17. **For** each cluster $C_l$ ($GP_b$ represents the best peak in $C_l$)
18.    **If** $LSP_b = 0$
19.       *MP* = $\{GP_i \in C_l \mid LSP_i = 1$ and $lsnum_i \geq 1\}$;
20.       *GP* = *GP* – *MP*;
21.       Expand the peak region of $GP_b$;
22.    **End If**
23. **End For**

Moreover, a local peak with a fitness very close to the global peaks may also be misclassified by PDM due to the very low value of *SFD*. According to our investigation, such a peak may be also located very close to the global peaks in the search space. Therefore, to deal with this problem, the mean-shift clustering method with the Gaussian kernel function is adopted to cluster the global peaks in *GP* (Step 16). In the mean-shift clustering, a parameter termed *bandwidth* is required. To simplify the notation, the best peak of a cluster $C_l$ refers to a peak $GP_b$ in *GP*. If $LSP_b = 0$, LSS finds the peaks in $C_l$ that have conducted at least one complete local search process and currently have an *LSP* value of 1, and then stores them in a set *MP* (Steps 17–19). Note that according to (12), a peak has a higher probability of obtaining *LSP*=0 if it has a better fitness, and thus LSS estimates that $GP_b$ meets the accuracy requirement and is a global peak. Those peaks with *LSP*=1 and *lsnum*≥1 in the same cluster as $GP_b$ are estimated as local peaks with a fitness very close to the global peak. Thus, the peaks in *MP* are removed from *GP* (Step 20). In addition, to prevent individuals from locating the removed peaks in the subsequent search, the peak region of $GP_b$ is expanded according to (15), which aims to cover all the peaks in *MP* (Step 21).

$$PD_b^d = \max_{GP_i \in MP}(PD_b^d, PD_i^d + |GP_b^d - GP_i^d|), d \in \{1, 2, ..., D\} \quad (15)$$

Note that the PDM has already obtained a great performance in distinguishing the found peaks according to experimental analysis in Section IV-D, and LSS further helps correct a few misclassifications.

## C. Landscape-Aware Reinitialization

According to PEP, an individual will locate a peak during its lifetime. After that, the individual will be reinitialized to start a new lifetime, which enhances the diversity to help locate more peaks. Random initialization is an intuitive approach. However, since individuals are more likely to locate a peak nearby, the initial position of the individual should be more efficient in helping an individual explore a new peak. To deal with this problem, PORDS and SDS are proposed to specify the reinitialization position. Overall, if the criterion of PORDS is met, the individual will be reinitialized according to PORDS. Otherwise, if the criterion of SDS is met, the individual will be reinitialized according to SDS. If the individual is not reinitialized by PORDS or SDS, random initialization is conducted. Details of PORDS and SDS are presented as follows.

*1) Potential Optimal Region Detection Strategy*

In the landscapes of some complex MMOPs, a global peak may be surrounded by multiple peaks, such a region is termed a potential optimal region. It is very difficult for an individual to explore and locate such a global peak. To deal with this challenge, PORDS is proposed, which aims to detect potential optimal regions and reinitialize individuals to explore these regions.

Specifically, for an individual that is not reinitialized in a potential optimal region in its current lifetime and locates a peak $P_k$ when its lifetime is exhausted, the mean-shift clustering with the Gaussian kernel function is used to cluster the peaks in $P$. In the cluster where $P_k$ belongs (termed as $C_{Pk}$), if the following three criteria are met, a potential optimal region is detected and the individual will be reinitialized in it.

  a) There are at least two peaks in $C_{Pk}$.
  b) *FGR* of the best peak in $C_{Pk}$ is less than 0.04.
  c) The *LSP* values of global peaks in $C_{Pk}$ are all equal to 1, and any of them has conducted at least a complete local search process.

If there is a potential optimal region, the individual is reinitialized inside it. Specifically, the individual is reinitialized in the average position of the peaks in $C_{Pk}$ (termed $P_{mean}$). Since such a region is usually quite small, thus the virtual individual range of the individual is initialized to a small value as (16). Moreover, simulated peak regions are not considered as taboo regions if an individual is reinitialized in a potential optimal region to help sufficiently explore a new global peak. If a new global peak has been successfully found in a potential optimal region, individuals will not be reinitialized in the same potential optimal region to save FEs.

$$R = \max\left(\left|\frac{P_{mean}^d - P_j^d}{2}\right|\right), P_j \in C_{Pk}, d \in \{1, 2, ..., D\} \quad (16)$$

*2) Subspace Division Strategy*

During the search process of LADE, a number of global peaks will gradually be found. The distribution of global peaks contains useful landscape knowledge that can provide search guidance to help locate more peaks. The purpose of SDS is to guide individuals to explore regions with fewer found global peaks for maintaining diversity. Since SDS requires a number of global peaks to capture the landscape, an indicator *SDP* is calculated by (17) and an individual will be reinitialized by SDS if *SDP* = 1. The probability of reinitializing an individual by SDS increases with the growing number of the found global peaks, which can enhance the effectiveness of *SDS*.

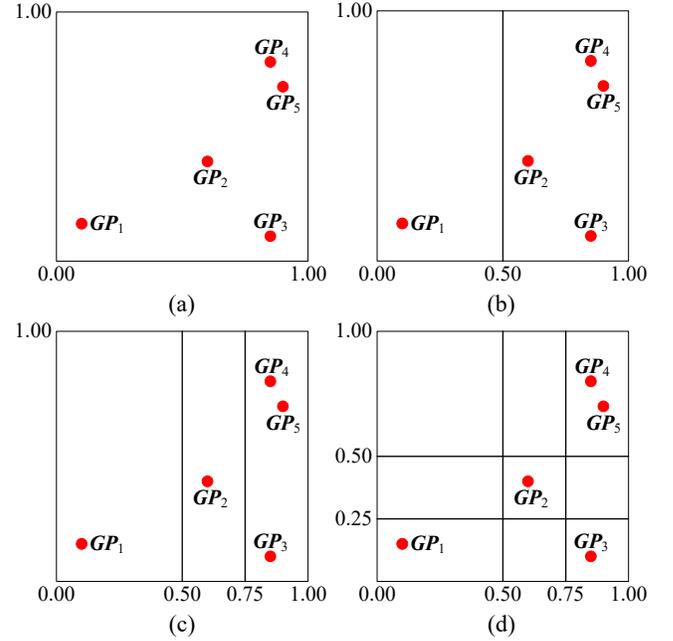

**Fig. 3.** Example of subspace division in a 2-D normalized search space. (a) Distribution of found global peaks in the search space. (b) Divide the horizontal dimension on 0.5. (c) Divide the horizontal dimension on 0.75. (d) Result of the subspace division.

$$SDP = \begin{cases} 1, & \text{if rand}(0,1) \leq \dfrac{1}{1+e^{20-|GP|}} \\ 0, & \text{otherwise} \end{cases} \quad (17)$$

The process of subspace division is described as follows. For each dimension, division is conducted on the midpoint of the corresponding dimension if the distribution of global peaks satisfies the following two criteria.

  a) In the corresponding dimension, there are global peaks distributed on both sides of the midpoint.
  b) In the corresponding dimension, the distance between the two farthest global peaks is greater than a quarter of the length of the subspace.

Based on the two criteria, the division process is conducted recursively in each dimension until no division can be further conducted. Take Fig. 3 as an example. In Fig. 3(a), there is a 2-D normalized search space and five global peaks, including $GP_1$ = (0.10, 0.15), $GP_2$ = (0.60, 0.40), $GP_3$ = (0.85, 0.10), $GP_4$ = (0.85, 0.80) and $GP_5$ = (0.90, 0.70). For the horizontal dimension, the current midpoint is 0.5, we can see that there are global peaks on both sides (i.e., $GP_1$ on the left side and the others on the right side). Then, the distance between the two farthest global peaks (i.e., $GP_1$ and $GP_5$) is 0.8 in the horizontal dimension, which is larger than a quarter of the length of the current subspace (i.e., 0.25). Therefore, SDS divides the search space on the corresponding midpoint of the horizontal dimension, as shown in Fig. 3(b). According to the same rule, the left subspace is not allowed for further division since there is no global peak on both sides of the corresponding midpoint (i.e., 0.25). The right subspace satisfies the division criteria and will be further divided on the corresponding midpoint (i.e., 0.75), as shown in Fig. 3(c). No subspace satisfies the division criteria in the horizontal dimension currently. Note that for the rightest subspace, the distance between the two farthest global peaks (i.e., $GP_3$ and $GP_5$) is 0.05, which is smaller than a quarter of the length of the current subspace (i.e., 0.0625). After that, SDS checks whether there are valid divisions in the vertical dimension recursively and finally obtains the result of subspace division shown in Fig. 3(d). We can see that nine subspaces of different scales are obtained.

If an individual is reinitialized by SDS (i.e., $SDP=1$), it selects a subspace according to the roulette wheel selection. $SSP_t$, which represents the probability of selecting the $t$-th subspace, is calculated by (18). $SN$ represents the total number of subspaces and $gpnum_t$ represents the number of global peaks in the $t$-th subspace. According to (18), an individual is more likely to be reinitialized in a subspace with fewer global peaks to maintain sufficient diversity. The individual is then reinitialized randomly in the selected subspace. The search of the individual during PEP is also restricted to the selected subspace. Specifically, $LB$ and $UB$ in (5) are set to the lower bound and the upper bound of the selected subspace respectively, and the offspring will be generated repeatedly until it is in the selected subspace. Note that if the minimal length of the selected subspace in all dimensions is less than 1/8, simulated peak regions are not considered as taboo regions during the current lifetime of the individual.

$$SSP_t = \frac{SN^{-gpnum_t}}{\sum_{sn=1}^{SN} SN^{-gpnum_{sn}}} \quad (18)$$

### D. Complete Procedure of LADE

The complete procedure of LADE is shown in Algorithm 4. At the beginning, an individual is randomly initialized (Step 1). Then, the individual explores the search space to locate a peak according to the PEP (Step 3). When its lifetime is exhausted, LADE distinguishes the detailed situation of the peak located by the individual according to PDM (Step 4) and refines the solution accuracy on the found global peaks according to LSS (Step 5). After that, LADE simulates the region of the peak located by the individual according to PRSS (Step 6). Lastly, LADE determines whether the individual is reinitialized by PORDS, SDS, or random initialization (Steps 7–13), and then the individual will start a new lifetime to locate more peaks. LADE iteratively conducts Steps 3–13 until the termination criterion is met, and finally outputs the results.

## IV. EXPERIMENTAL STUDIES

### A. Benchmark Functions and Performance Metrics

The MMOP benchmark set presented in [41], which is widely used recently [15]−[19], is adopted to evaluate the performance of LADE. There are 20 benchmark functions in the adopted benchmark set, and all of them are maximization problems. Among them, $F_1$–$F_5$ are simple functions; $F_6$–$F_{10}$ are scalable functions with a large number of global peaks; $F_{11}$–$F_{20}$ are composition functions constructed from different basic functions. More information about the benchmark functions can be found in Table S.I in the supplementary material.

---

**Algorithm 4** LADE
1. Randomly initialize an individual $X$;
2. **While** the termination criterion is not met
3.    Conduct PEP;
4.    Conduct PDM;
5.    Conduct LSS;
6.    Conduct PRSS;
7.    Conduct PORDS;
8.    **If** $X$ is not reinitialized by PORDS
9.      Conduct SDS;
10.     **If** $X$ is not reinitialized by SDS
11.       Randomly reinitialize $X$;
12.     **End If**
13.    **End If**
14. **End While**

---

Peak ratio (PR) and success rate (SR) are used to evaluate the performance of multimodal optimization algorithms over multiple runs. Given an accuracy level $\varepsilon$ and a maximum number of function evaluations, the algorithm is independently run 50 times on each benchmark function, and the global peaks found in each run are recorded. PR is calculated by (19), which denotes the average percentage of global peaks found over multiple runs. $NPF_i$ represents the number of global peaks found in the $i$-th run. $NKP$ represents the number of known global peaks and $NR$ represents the total number of runs.

$$PR = \frac{\sum_{i=1}^{NR} NPF_i}{NKP \times NR} \quad (19)$$

SR is calculated by (20), which represents the percentage of the runs in which all global peaks are found. $NSR$ represents the number of runs where all global peaks are successfully found.

$$SR = \frac{NSR}{NR} \quad (20)$$

According to the suggestion of the benchmark set [41], three accuracy levels (i.e., $\varepsilon = 1.0\text{E}{-}03$, $\varepsilon = 1.0\text{E}{-}04$, and $\varepsilon = 1.0\text{E}{-}05$) are adopted in the experiments. We mainly discuss the experimental results at the highest accuracy level $\varepsilon = 1.0\text{E}{-}05$, while the results at the other two accuracy levels will be presented in the supplementary material. The maximum number of function evaluations on each function is also set according to [41].

Parameters in LADE are set as follows: $\lambda$ in PDM is set to $10^{-2}$; *bandwidth* in the mean-shift clustering is set to 0.1; $\sigma_{ini}$, $\sigma_{ter}$, and $dt$ in LSS are set to $10^{-4}$, $10^{-11}$, and 40, respectively.

TABLE I
EXPERIMENTAL RESULTS OF PR AND SR AT THE ACCURACY LEVEL $\varepsilon$ =1.0E–05

| Func | LADE | | DIDE | | FBK-DE | | NBNC-PSO-ES | | PMODE | | ESPDE | |
|---|---|---|---|---|---|---|---|---|---|---|---|---|
| | PR | SR | PR | SR | PR | SR | PR | SR | PR | SR | PR | SR |
| $F_1$ | **1.000** | 1.000 | **1.000**(=) | 1.000 | **1.000**(=) | 1.000 | **1.000**(=) | 1.000 | **1.000**(=) | 1.000 | **1.000**(=) | 1.000 |
| $F_2$ | **1.000** | 1.000 | **1.000**(=) | 1.000 | **1.000**(=) | 1.000 | **1.000**(=) | 1.000 | **1.000**(=) | 1.000 | **1.000**(=) | 1.000 |
| $F_3$ | **1.000** | 1.000 | **1.000**(=) | 1.000 | **1.000**(=) | 1.000 | **1.000**(=) | 1.000 | **1.000**(=) | 1.000 | **1.000**(=) | 1.000 |
| $F_4$ | **1.000** | 1.000 | **1.000**(=) | 1.000 | **1.000**(=) | 1.000 | **1.000**(=) | 1.000 | **1.000**(=) | 1.000 | **1.000**(=) | 1.000 |
| $F_5$ | **1.000** | 1.000 | **1.000**(=) | 1.000 | **1.000**(=) | 1.000 | **1.000**(=) | 1.000 | **1.000**(=) | 1.000 | **1.000**(=) | 1.000 |
| $F_6$ | **1.000** | 1.000 | **1.000**(=) | 1.000 | 0.990(+) | 0.820 | **1.000**(=) | 1.000 | **1.000**(=) | 1.000 | 0.998(+) | 0.960 |
| $F_7$ | **1.000** | 1.000 | 0.920(+) | 0.020 | 0.813(+) | 0.000 | 0.965(+) | 0.120 | 0.672(+) | 0.000 | 0.963(+) | 0.360 |
| $F_8$ | **1.000** | 1.000 | 0.689(+) | 0.000 | 0.823(+) | 0.000 | 0.807(+) | 0.000 | 0.616(+) | 0.000 | 0.856(+) | 0.000 |
| $F_9$ | **1.000** | 1.000 | 0.561(+) | 0.000 | 0.425(+) | 0.000 | 0.540(+) | 0.000 | 0.324(+) | 0.000 | 0.727(+) | 0.000 |
| $F_{10}$ | **1.000** | 1.000 | **1.000**(=) | 1.000 | **1.000**(=) | 1.000 | **1.000**(=) | 1.000 | **1.000**(=) | 1.000 | **1.000**(=) | 1.000 |
| $F_{11}$ | **1.000** | 1.000 | **1.000**(=) | 1.000 | **1.000**(=) | 1.000 | **1.000**(=) | 1.000 | **1.000**(=) | 1.000 | **1.000**(=) | 1.000 |
| $F_{12}$ | **1.000** | 1.000 | **1.000**(=) | 1.000 | 0.935(+) | 0.480 | **1.000**(=) | 1.000 | **1.000**(=) | 1.000 | 0.840(+) | 0.200 |
| $F_{13}$ | **1.000** | 1.000 | 0.957(+) | 0.740 | **1.000**(=) | 1.000 | **1.000**(=) | 1.000 | 0.953(+) | 0.720 | 0.773(+) | 0.080 |
| $F_{14}$ | **0.897** | 0.400 | 0.733(+) | 0.000 | 0.890(+) | 0.360 | 0.847(+) | 0.100 | 0.800(+) | 0.000 | 0.713(+) | 0.000 |
| $F_{15}$ | **0.750** | 0.000 | 0.748(+) | 0.000 | 0.728(+) | 0.000 | 0.733(+) | 0.000 | **0.750**(=) | 0.000 | 0.730(+) | 0.000 |
| $F_{16}$ | 0.670 | 0.000 | 0.667(+) | 0.000 | 0.707(−) | 0.000 | 0.723(−) | 0.000 | 0.667(+) | 0.000 | 0.667(+) | 0.000 |
| $F_{17}$ | **0.750** | 0.000 | 0.588(+) | 0.000 | 0.630(+) | 0.000 | 0.715(+) | 0.000 | 0.405(+) | 0.000 | 0.680(+) | 0.000 |
| $F_{18}$ | **0.667** | 0.000 | **0.667**(=) | 0.000 | **0.667**(=) | 0.000 | **0.667**(=) | 0.000 | 0.500(+) | 0.000 | 0.660(+) | 0.000 |
| $F_{19}$ | 0.578 | 0.000 | 0.535(+) | 0.000 | 0.518(+) | 0.000 | 0.538(+) | 0.000 | 0.245(+) | 0.000 | 0.440(+) | 0.000 |
| $F_{20}$ | 0.383 | 0.000 | 0.345(+) | 0.000 | 0.445(−) | 0.000 | 0.483(−) | 0.000 | 0.240(+) | 0.000 | 0.095(+) | 0.000 |
| + (LADE is better) | | | 10 | | 9 | | 7 | | 10 | | 13 | |
| = (Equal) | | | 10 | | 9 | | 11 | | 10 | | 7 | |
| − (LADE is worse) | | | 0 | | 2 | | 2 | | 0 | | 0 | |
| Func | NDC-DE | | NetCDE$_{MMOPs}$ | | NEA2 | | NMMSO | | RS-CMSA-ES | | RS-CMSA-ESII | |
| | PR | SR | PR | SR | PR | SR | PR | SR | PR | SR | PR | SR |
| $F_1$ | **1.000**(=) | 1.000 | **1.000**(=) | 1.000 | **1.000**(=) | 1.000 | **1.000**(=) | 1.000 | **1.000**(=) | 1.000 | **1.000**(=) | 1.000 |
| $F_2$ | **1.000**(=) | 1.000 | **1.000**(=) | 1.000 | **1.000**(=) | 1.000 | **1.000**(=) | 1.000 | **1.000**(=) | 1.000 | **1.000**(=) | 1.000 |
| $F_3$ | **1.000**(=) | 1.000 | **1.000**(=) | 1.000 | **1.000**(=) | 1.000 | **1.000**(=) | 1.000 | **1.000**(=) | 1.000 | **1.000**(=) | 1.000 |
| $F_4$ | **1.000**(=) | 1.000 | **1.000**(=) | 1.000 | 0.990(+) | 0.960 | **1.000**(=) | 1.000 | **1.000**(=) | 1.000 | **1.000**(=) | 1.000 |
| $F_5$ | **1.000**(=) | 1.000 | **1.000**(=) | 1.000 | **1.000**(=) | 1.000 | **1.000**(=) | 1.000 | **1.000**(=) | 1.000 | **1.000**(=) | 1.000 |
| $F_6$ | 0.999(+) | 0.980 | **1.000**(=) | 1.000 | 0.950(+) | 0.380 | 0.997(+) | 0.940 | 0.999(+) | 0.980 | **1.000**(=) | 1.000 |
| $F_7$ | 0.881(+) | 0.000 | 0.971(+) | 0.333 | 0.911(+) | 0.040 | **1.000**(=) | 1.000 | 0.998(+) | 0.920 | **1.000**(=) | 1.000 |
| $F_8$ | 0.935(+) | 0.000 | 0.999(+) | 0.902 | 0.239(+) | 0.000 | 0.980(+) | 0.180 | 0.875(+) | 0.000 | 0.997(+) | 0.760 |
| $F_9$ | 0.458(+) | 0.000 | 0.513(+) | 0.000 | 0.579(+) | 0.000 | 0.913(+) | 0.000 | 0.734(+) | 0.000 | 0.990(+) | 0.140 |
| $F_{10}$ | **1.000**(=) | 1.000 | **1.000**(=) | 1.000 | 0.980(+) | 0.760 | **1.000**(=) | 1.000 | **1.000**(=) | 1.000 | **1.000**(=) | 1.000 |
| $F_{11}$ | **1.000**(=) | 1.000 | 0.977(+) | 0.863 | 0.960(+) | 0.760 | **1.000**(=) | 1.000 | 0.997(+) | 0.980 | **1.000**(=) | 1.000 |
| $F_{12}$ | 0.941(+) | 0.520 | 0.914(+) | 0.530 | 0.833(+) | 0.140 | 0.998(+) | 0.980 | 0.948(+) | 0.580 | **1.000**(=) | 1.000 |
| $F_{13}$ | **1.000**(=) | 1.000 | 0.667(+) | 0.000 | 0.947(+) | 0.700 | 0.990(+) | 0.940 | 0.997(+) | 0.980 | 0.993(+) | 0.960 |
| $F_{14}$ | 0.743(+) | 0.100 | 0.670(+) | 0.000 | 0.800(+) | 0.060 | 0.703(+) | 0.000 | 0.803(+) | 0.100 | 0.850(+) | 0.100 |
| $F_{15}$ | 0.720(+) | 0.000 | 0.630(+) | 0.000 | 0.713(+) | 0.000 | 0.668(+) | 0.000 | 0.745(+) | 0.000 | **0.750**(=) | 0.000 |
| $F_{16}$ | 0.667(+) | 0.000 | 0.667(+) | 0.000 | 0.673(−) | 0.000 | 0.660(+) | 0.000 | 0.667(+) | 0.000 | **0.833**(−) | 0.000 |
| $F_{17}$ | 0.633(+) | 0.000 | 0.480(+) | 0.000 | 0.695(+) | 0.000 | 0.538(+) | 0.000 | 0.695(+) | 0.000 | **0.750**(=) | 0.000 |
| $F_{18}$ | **0.667**(=) | 0.000 | **0.667**(=) | 0.000 | 0.663(+) | 0.000 | 0.633(+) | 0.000 | **0.667**(=) | 0.000 | **0.667**(=) | 0.000 |
| $F_{19}$ | 0.410(+) | 0.000 | 0.461(+) | 0.000 | 0.667(−) | 0.000 | 0.443(+) | 0.000 | 0.508(+) | 0.000 | **0.703**(−) | 0.000 |
| $F_{20}$ | 0.345(+) | 0.000 | 0.380(+) | 0.000 | 0.350(+) | 0.000 | 0.178(+) | 0.000 | 0.468(−) | 0.000 | **0.618**(−) | 0.000 |
| + | | | 11 | | 12 | | 14 | | 12 | | 12 | | 4 |
| = | | | 9 | | 8 | | 4 | | 8 | | 7 | | 13 |
| − | | | 0 | | 0 | | 2 | | 0 | | 1 | | 3 |

### B. Comparison with State-of-the-Art Multimodal Optimization Algorithms

The proposed LADE is compared with the following 11 state-of-the-art multimodal optimization algorithms:

1) Multimodal optimization algorithms published in well-known journals in the last four years, including DIDE [13], FBK-DE [32], NBNC-PSO-ES [10], PMODE [16], ESPDE [31], NDC-DE [33], and NetCDE$_{MMOPs}$ [17].
2) The winners of the IEEE CEC competitions for multimodal optimization in recent years, including NEA2 [28], NMMSO [30], RS-CMSA-ES [39], RS-CMSA-ESII [15]. Note that RS-CMSA-ESII is the winner of the latest competition.

The experimental results of the compared algorithms are referred to their corresponding references or their corresponding results in IEEE CEC competitions for multimodal optimization.

Table I shows the experimental results of PR and SR of LADE and the compared algorithms at the accuracy level $\varepsilon$ = 1.0E–05. The symbols "+", "=", and "−" represent that LADE is better than, equal to, and worse than the compared algorithm, respectively. The best PR results among all algorithms on each function are highlighted in bold. Detailed discussions of the results in Table I are as follows.

TABLE II
EXPERIMENTAL RESULTS OF PR AND SR IN THE COMPONENT ANALYSIS AT THE ACCURACY LEVEL $\varepsilon =1.0E-05$

| Func | LADE | | NPR* | | NPO* | | NSD* | | NPS* | |
|---|---|---|---|---|---|---|---|---|---|---|
| | PR | SR | PR | SR | PR | SR | PR | SR | PR | SR |
| $F_1$ | **1.000** | 1.000 | **1.000**(=) | 1.000 | **1.000**(=) | 1.000 | **1.000**(=) | 1.000 | **1.000**(=) | 1.000 |
| $F_2$ | **1.000** | 1.000 | **1.000**(=) | 1.000 | **1.000**(=) | 1.000 | **1.000**(=) | 1.000 | **1.000**(=) | 1.000 |
| $F_3$ | **1.000** | 1.000 | **1.000**(=) | 1.000 | **1.000**(=) | 1.000 | **1.000**(=) | 1.000 | **1.000**(=) | 1.000 |
| $F_4$ | **1.000** | 1.000 | **1.000**(=) | 1.000 | **1.000**(=) | 1.000 | **1.000**(=) | 1.000 | **1.000**(=) | 1.000 |
| $F_5$ | **1.000** | 1.000 | **1.000**(=) | 1.000 | **1.000**(=) | 1.000 | **1.000**(=) | 1.000 | **1.000**(=) | 1.000 |
| $F_6$ | **1.000** | 1.000 | **1.000**(=) | 1.000 | **1.000**(=) | 1.000 | **1.000**(=) | 1.000 | **1.000**(=) | 1.000 |
| $F_7$ | **1.000** | 1.000 | **1.000**(=) | 1.000 | **1.000**(=) | 1.000 | 0.999(+) | 0.960 | 0.998(+) | 0.940 |
| $F_8$ | **1.000** | 1.000 | 0.961(+) | 0.040 | **1.000**(=) | 1.000 | 0.994(+) | 0.580 | 0.988(+) | 0.360 |
| $F_9$ | **1.000** | 1.000 | **1.000**(=) | 1.000 | **1.000**(=) | 1.000 | 0.714(+) | 0.000 | 0.711(+) | 0.000 |
| $F_{10}$ | **1.000** | 1.000 | **1.000**(=) | 1.000 | **1.000**(=) | 1.000 | **1.000**(=) | 1.000 | **1.000**(=) | 1.000 |
| $F_{11}$ | **1.000** | 1.000 | **1.000**(=) | 1.000 | **1.000**(=) | 1.000 | **1.000**(=) | 1.000 | **1.000**(=) | 1.000 |
| $F_{12}$ | **1.000** | 1.000 | 0.885(+) | 0.320 | **1.000**(=) | 1.000 | **1.000**(=) | 1.000 | **1.000**(=) | 1.000 |
| $F_{13}$ | **1.000** | 1.000 | **1.000**(=) | 1.000 | 0.997(+) | 0.980 | 0.997(+) | 0.980 | 0.997(+) | 0.980 |
| $F_{14}$ | 0.897 | 0.400 | **0.923**(−) | 0.580 | 0.850(+) | 0.260 | 0.887(+) | 0.360 | 0.843(+) | 0.160 |
| $F_{15}$ | **0.750** | 0.000 | 0.688(+) | 0.000 | 0.668(+) | 0.000 | **0.750**(=) | 0.000 | 0.695(+) | 0.000 |
| $F_{16}$ | **0.670** | 0.000 | 0.667(+) | 0.000 | 0.667(+) | 0.000 | 0.667(+) | 0.000 | 0.667(+) | 0.000 |
| $F_{17}$ | **0.750** | 0.000 | 0.580(+) | 0.000 | 0.548(+) | 0.000 | **0.750**(=) | 0.000 | 0.590(+) | 0.000 |
| $F_{18}$ | **0.667** | 0.000 | 0.640(+) | 0.000 | **0.667**(=) | 0.000 | **0.667**(=) | 0.000 | **0.667**(=) | 0.000 |
| $F_{19}$ | **0.578** | 0.000 | 0.540(+) | 0.000 | 0.465(+) | 0.000 | **0.578**(=) | 0.000 | 0.445(+) | 0.000 |
| $F_{20}$ | **0.383** | 0.000 | 0.328(+) | 0.000 | 0.268(+) | 0.000 | 0.360(+) | 0.000 | 0.298(+) | 0.000 |
| + (LADE is better) | | | 8 | | 7 | | 7 | | 10 | |
| = (Equal) | | | 11 | | 13 | | 13 | | 10 | |
| − (LADE is worse) | | | 1 | | 0 | | 0 | | 0 | |

1) LADE obtains the best performance on 17 functions, and can stably find all global peaks on 13 functions (SR=1.000). These two numbers are both the best compared with other algorithms. In addition, LADE outperforms 10 compared algorithms on at least 7 functions; 6 compared algorithms cannot outperform LADE on any function.

2) For 15 low-dimensional functions $F_1$−$F_{15}$ ($D \leq 3$), LADE obtains the best results on all of them. LADE is the only algorithm that can stably find all global peaks on $F_1$−$F_{13}$. On $F_8$, $F_9$ and $F_{14}$, LADE outperforms all the compared algorithms. Particularly on $F_9$ which has 216 global peaks, the PR result of LADE is at least 0.4 higher than 7 compared algorithms. Compared with RS-CMSA-ESII which obtains a PR of 0.990 on $F_9$, LADE can still find 2 more peaks on average.

3) For 5 high-dimensional functions $F_{16}$−$F_{20}$ ($D \geq 5$), LADE performs the best on $F_{17}$ and $F_{18}$. On $F_{16}$, $F_{19}$, and $F_{20}$, LADE performs better than 7, 9, and 7 compared algorithms, respectively. RS-CMSA-ESII, the winner of the latest IEEE CEC competition, shows a competitive advantage on high-dimensional functions, particularly on $F_{20}$.

These results show that in low-dimensional search space, landscape knowledge can be obtained and accumulated by LADE efficiently, helping achieve good performance. The search space expands exponentially with the increase of dimension size. Thus, the search trajectories of LADE are sparse in solving high-dimensional functions, increasing the difficulty of acquiring landscape knowledge.

The experiment results at the other two accuracy levels ($\varepsilon = 1.0E-03$ and $\varepsilon = 1.0E-04$) are presented in Tables S.II and S.III in the supplementary material, respectively. The experiment results of LADE are identical at the three accuracy levels, while the performance of the majority of compared algorithms decreases as the accuracy level increases, which validates the effectiveness of LADE in refining the solution accuracy.

*C. Component Analysis*

In this section, the effect of three important components in LADE (i.e., PRSS, PORDS, and SDS) are discussed as follows. The experimental results at the accuracy level $\varepsilon = 1.0E-05$ are shown in Table II with the best PR results on each function highlighted in bold. The experimental results at the other two accuracy levels are presented in Tables S.IV and S.V in the supplementary material, respectively. To simplify the notation, the LADE variants without PRSS, PORDS, SDS, both PORDS and SDS are abbreviated as NPR*, NPO*, NSD* and NPS*, respectively. Note that in NPS*, only random reinitialization is adopted.

LADE outperforms NPR* on 8 functions, validating the effectiveness of PRSS. PRSS keeps an individual away from the regions of the found peaks and can encourage the individual to explore new peaks. Moreover, the 8 functions on which LADE has better performance all have a lot of local peaks, showing that PRSS can avoid an individual repeatedly locating the found local peaks and save FEs to help locate more global peaks.

Compared with NPO*, LADE obtains better performance on 7 functions, particularly on $F_{17}$, $F_{19}$ and $F_{20}$. $F_{17}$, $F_{19}$ and $F_{20}$ are complex composition functions with relatively high dimension sizes in the benchmark set. To efficiently deal with the complex landscape, PORDS detects the potential optimal region where a global peak may exist and is surrounded by multiple found peaks. Then, PORDS lets an individual explore such regions to help locate a new global peak. As another reinitialization strategy, SDS divides the search space into a group of subspaces according to the distribution of the found global peaks, and adaptively reinitializes an individual in one of the subspaces. A subspace with fewer global peaks is preferred. Comparing LADE and NSD*, the effectiveness of SDS is significant on the functions with a large number of global peaks, particularly on $F_9$ in which there are 216 global peaks. Compared with NPS*, LADE outperforms it on 10 functions, validating the effectiveness of landscape-aware reinitialization in LADE.

TABLE III
EXPERIMENTAL RESULTS OF PR IN THE PARAMETER ANALYSIS OF $\lambda$ AT THE ACCURACY LEVEL $\varepsilon = 1.0\text{E}{-}05$

| Func | $\lambda = 10^{-2}$ | $\lambda = 10^{-4}$ | $\lambda = 10^{-3}$ | $\lambda = 10^{-1}$ | $\lambda = 1$ |
|---|---|---|---|---|---|
| $F_1$ | **1.000** | **1.000**(=) | **1.000**(=) | **1.000**(=) | **1.000**(=) |
| $F_2$ | **1.000** | **1.000**(=) | **1.000**(=) | **1.000**(=) | **1.000**(=) |
| $F_3$ | **1.000** | **1.000**(=) | **1.000**(=) | **1.000**(=) | **1.000**(=) |
| $F_4$ | **1.000** | **1.000**(=) | **1.000**(=) | **1.000**(=) | 0.935(+) |
| $F_5$ | **1.000** | **1.000**(=) | **1.000**(=) | **1.000**(=) | **1.000**(=) |
| $F_6$ | **1.000** | **1.000**(=) | **1.000**(=) | 0.994(+) | 0.941(+) |
| $F_7$ | **1.000** | **1.000**(=) | **1.000**(=) | **1.000**(=) | 0.990(+) |
| $F_8$ | **1.000** | 0.685(+) | 0.991(+) | **1.000**(=) | 0.914(+) |
| $F_9$ | **1.000** | **1.000**(=) | **1.000**(=) | **1.000**(=) | 0.991(+) |
| $F_{10}$ | **1.000** | **1.000**(=) | **1.000**(=) | 0.998(+) | 0.952(+) |
| $F_{11}$ | **1.000** | 0.977(+) | 0.997(+) | **1.000**(=) | 0.683(+) |
| $F_{12}$ | **1.000** | 0.903(+) | 0.998(+) | 0.998(+) | 0.798(+) |
| $F_{13}$ | **1.000** | 0.773(+) | 0.997(+) | 0.943(+) | 0.670(+) |
| $F_{14}$ | **0.897** | 0.683(+) | 0.760(+) | 0.760(+) | 0.667(+) |
| $F_{15}$ | **0.750** | 0.460(+) | 0.518(+) | **0.750**(=) | 0.748(+) |
| $F_{16}$ | **0.670** | 0.607(+) | 0.667(+) | 0.667(+) | 0.613(+) |
| $F_{17}$ | **0.750** | 0.310(+) | 0.293(+) | 0.740(+) | 0.715(+) |
| $F_{18}$ | **0.667** | 0.503(+) | **0.667**(=) | 0.617(+) | 0.347(+) |
| $F_{19}$ | **0.578** | 0.128(+) | 0.463(+) | 0.543(+) | 0.438(+) |
| $F_{20}$ | **0.383** | 0.288(+) | 0.378(+) | 0.265(+) | 0.238(+) |
| + | | 11 | 10 | 10 | 16 |
| = | | 9 | 10 | 10 | 4 |
| − | | 0 | 0 | 0 | 0 |

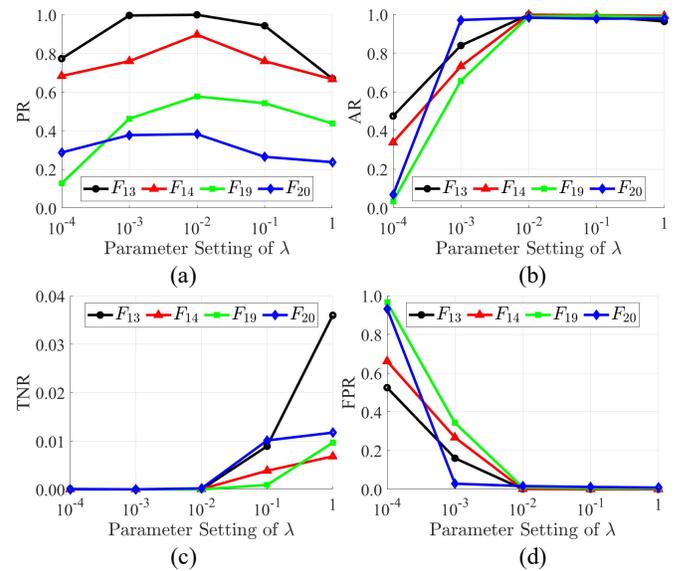

**Fig. 4.** Experimental results of (a) PR, (b) AR, (c) TNR, and (d) FPR for five different settings of $\lambda$ on four functions $F_{13}$, $F_{14}$, $F_{19}$, and $F_{20}$ at the accuracy level $\varepsilon = 1.0\text{E}{-}05$.

In conclusion, PRSS and SDS maintain sufficient diversity to help locate more peaks, while PORDS provides efficient search guidance to help an individual explore a global peak in a complex landscape. The PRSS, PORDS, and SDS cooperatively help LADE obtain an overall good performance.

*D. Parameter Analysis*

$\lambda$ in PDM is a very important parameter, since most of the components in LADE rely on an accurate distinction of the found peaks. To investigate the influence of $\lambda$, four settings of $\lambda$ (i.e., $10^{-4}$, $10^{-3}$, $10^{-1}$, and 1) are tested and compared with the adopted setting (i.e., $10^{-2}$) in LADE. The experimental results of PR at the accuracy level $\varepsilon = 1.0\text{E}{-}05$ are presented in Table III. The best results on each function are highlighted in bold. The experimental results at the other two accuracy levels are presented in Tables S.VI and S.VII in the supplementary material, respectively.

According to (7), a smaller value of $\lambda$ results in a smaller value of *SFD*, and thus a peak is more likely to be classified as a global peak, which can avoid misclassifying a global peak as a local peak. However, local peaks may be more likely to be misclassified as global peaks at the same time. We can see that on some composition functions with numerous local peaks, two settings of small values (i.e., $10^{-4}$ and $10^{-3}$) obtain unsatisfying PR results, particularly on $F_{15}$ and $F_{17}$. Oppositely, a large value of $\lambda$ can avoid misclassifying local peaks as global peaks, but some global peaks may be misclassified as local peaks. Therefore, the settings of $10^{-1}$ and 1 do not perform well, particularly on $F_{18}$–$F_{20}$. Overall, the adopted setting in LADE can balance the misclassification of local peaks and global peaks, obtaining the best performance.

To make a detailed investigation, each distinction in PDM is recorded and three metrics are calculated, i.e., accuracy rate (AR), true negative rate (TNR), and false positive rate (FPR). AR represents the percentage of correct distinctions over multiple runs; TNR represents the percentage of distinctions that misclassify global peaks over multiple runs; FPR represents the percentage of distinctions that misclassify local peaks over multiple runs.

The experimental results of PR, AR, TNR, and FPR for five different settings of $\lambda$ on four representative functions (i.e., $F_{13}$, $F_{14}$, $F_{19}$, and $F_{20}$) at the accuracy level $\varepsilon = 1.0\text{E}{-}05$ are shown in Fig. 4. We can see that the larger the $\lambda$, the larger the TNR, which shows that a larger $\lambda$ will misclassify more global peaks as local peaks. In addition, the smaller the $\lambda$, the larger the FPR, which shows that a smaller $\lambda$ will misclassify more local peaks as global peaks. In this case, FEs are wasted in refining solution accuracy on a lot of local peaks. The adopted setting exhibits the overall highest AR, which provides the most accurate distinction and helps LADE obtain the best PR results.

## V. CONCLUSION

In this paper, a new LADE algorithm is proposed for MMOPs, which utilizes landscape knowledge to maintain sufficient diversity and provide efficient search guidance. First, the landscape-aware peak exploration allows each individual to evolve adaptively to locate a peak. To avoid repeatedly locating the found peaks, an individual is guided to keep away from the regions of found peaks simulated according to search trajectories. Second, the landscape-aware peak distinction is designed to distinguish whether a found peak is a global or local peak and refine the solution accuracy on global peaks. Third, the landscape-aware reinitialization determines the initial position of an individual according to the distribution and distinction of found peaks, which is beneficial to enhance search efficiency and help locate more peaks. The widely-used benchmark set with 20 MMOPs is adopted in the experiment. Experimental results show that the overall performance of LADE is better or competitive compared with seven well-performed algorithms proposed recently and four winner algorithms in the IEEE CEC competitions for multimodal optimization. In the future, we will extend LADE to solve more complex MMOPs, such as dynamic MMOPs [42].